\documentclass[letterpaper, 10 pt, conference]{ieeeconf}  % Comment this line out if you need a4paper

\IEEEoverridecommandlockouts                              % This command is only needed if 
\usepackage{amssymb}
\usepackage{graphicx}
\usepackage{tabu}
\usepackage{wrapfig}
\usepackage{xcolor}
\usepackage{amsmath}
\usepackage{flushend}
\usepackage{tikz}

\usepackage{booktabs}
\usepackage{cleveref}

\newcommand{\stopgrad}{|_{\nabla_{\theta}}}

\newcommand{\LLDbP}{LL-DbP }
\newcommand{\MDbP}{M-DbP }

\title{\LARGE \bf
Depth by Poking: Learning to Estimate Depth from Self-Supervised Grasping
}

\author{Ben Goodrich$^{*,1}$, Alex Kuefler$^{*}$, William D. Richards$^{1}$
\thanks{*Equal contribution, $^{1}$Osaro Inc.,  \{ben, will\}@osaro.com}
}

\begin{document}

\maketitle
\thispagestyle{empty}
\pagestyle{empty}

%%%%%%%%%%%%%%%%%%%%%%%%%%%%%%%%%%%%%%%%%%%%%%%%%%%%%%%%%%%%%%%%%%%%%%%%%%%%%%%%
\begin{abstract}
Accurate depth estimation remains an open problem for robotic manipulation; even state of the art techniques including structured light and LiDAR sensors fail on reflective or transparent surfaces. We address this problem by training a neural network model to estimate depth from RGB-D images, using labels from physical interactions between a robot and its environment. Our network predicts, for each pixel in an input image, the z position that a robot's end effector would reach if it attempted to grasp or poke at the corresponding position. Given an autonomous grasping policy, our approach is self-supervised as end effector position labels can be recovered through forward kinematics, without human annotation. Although gathering such physical interaction data is expensive, it is necessary for training and routine operation of state of the art manipulation systems. Therefore, this depth estimator comes ``for free'' while collecting data for other tasks (e.g., grasping, pushing, placing). We show our approach achieves significantly lower root mean squared error than traditional structured light sensors and unsupervised deep learning methods on difficult, industry-scale jumbled bin datasets.
\end{abstract}

%%%%%%%%%%%%%%%%%%%%%%%%%%%%%%%%%%%%%%%%%%%%%%%%%%%%%%%%%%%%%%%%%%%%%%%%%%%%%%%%
\section{INTRODUCTION}

Estimating the distance of objects to a given viewpoint is a long-standing problem in computer vision with implications for robotics. In the context of learning-based robot grasping and manipulation, reliable depth estimates are important for scene understanding, motion planning, and control. Depth maps provide information about a visual scene that is invariant to color. Data augmentation can also impose model invariance to nuisance features, but depth imagery provides additional information useful for robotic manipulation \cite{pmlr-v78-viereck17a, morrison2018closing, Mahler2018DexNet3C}. During inference, knowing the depth of candidate grasp or placement points allows a gripper to navigate into position quickly, without relying on sensorimotor feedback.

Depth maps are often obtained with specialized hardware. Due to their commercial availability and ease of use, structured light sensors are particularly popular for robotics. Light detection and ranging (LiDAR) is also common for autonomous driving, as structured light approaches encounter difficulty in outdoor settings \cite{geiger2012we}. Both approaches suffer from sensor noise, particularly when encountering shiny, reflective, transparent, or textureless surfaces  \cite{tatoglu2012point, yang2011solving, ryan2016hyperdepth, lysenkov2013recognition}. In large-scale pick-and-place contexts (e.g., e-commerce, automated warehousing) the types of items to be manipulated may be diverse, unpredictable, or in some cases, entirely unknown a priori. Therefore, it is important that depth estimation approaches generalize to a wide range of surfaces.

In this work, we propose Depth by Poking (DbP): A depth estimation technique for robotic grasping, trained using only self-supervision. Our method uses robot pose and grasp success as training signals that are not available in general computer vision settings, but are readily obtained in pick-and-place environments. 

We formulate depth estimation as an image-to-image translation problem \cite{isola2017image} where RGB or noisy depth images are translated into accurate depth maps by a deep fully-convolutional network (FCN). Our model is trained to produce depth maps by performing per-pixel regression where the robot effector's pose during a grasp provides a training label. In this setting, data are sparsely labeled as only a single pixel (the grasp point) per training image has a ground truth depth value. But we show that DbP generalizes over complete depth maps given enough training data and argue that the burden of data collection is small in the context of existing, self-supervised grasping approaches.

\begin{figure}[t!]
  \label{fig:train_example}
  \centering
  \includegraphics[width=0.45\textwidth]{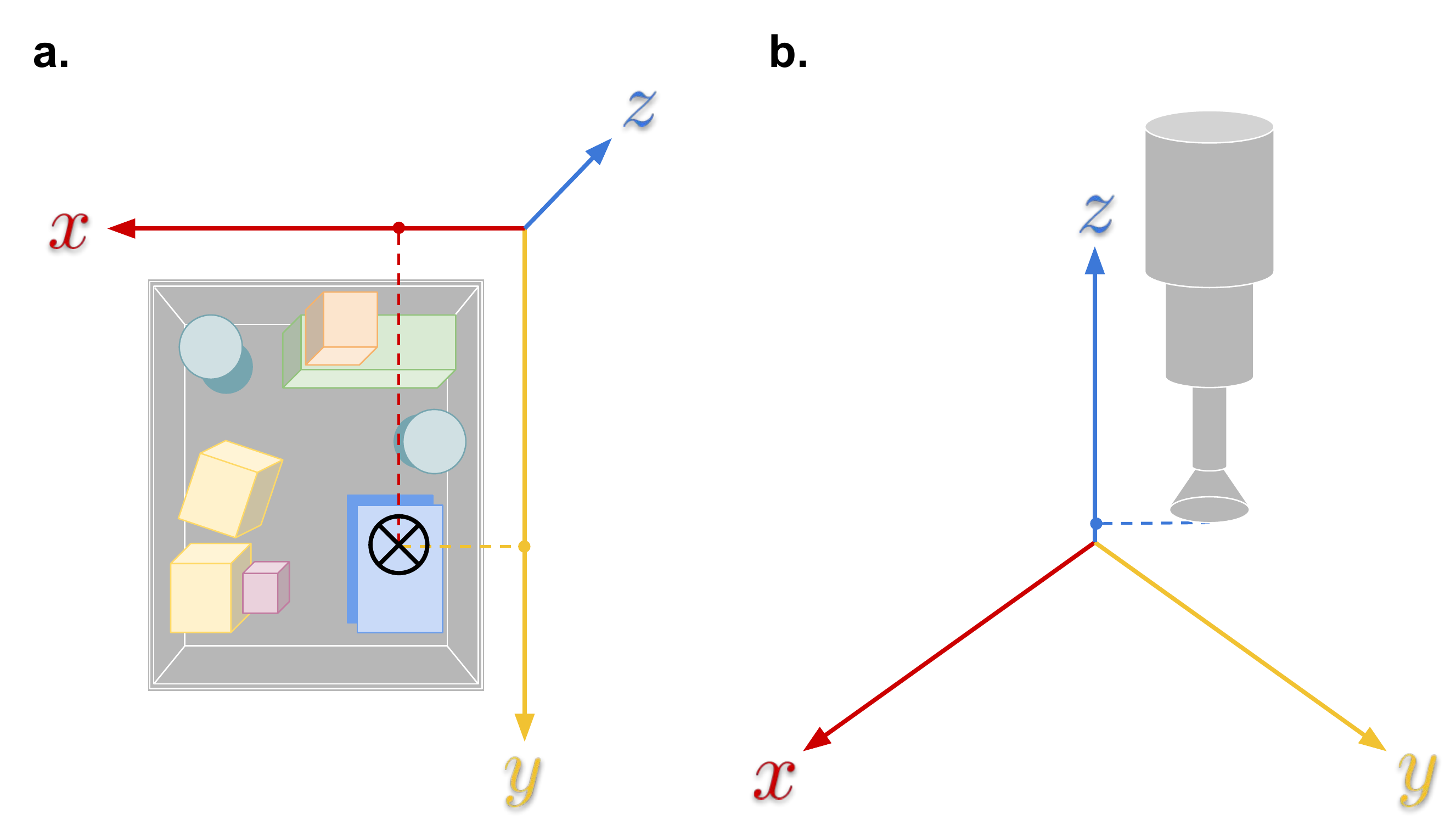}
  \caption{Training data is gathered by attempting picks. Collected samples consist of \textbf{a.} top-down images of a cluttered bin and a grasp point projected into pixel coordinates (black ``x''), and \textbf{b.} label position of tooltip along the $z$-axis.}
\end{figure}

%===============================================================================

\section{RELATED WORK}

\textbf{Depth Estimation.} There is large body of work on estimating the distance of objects to a given viewpoint. Current state of the art approaches combine stereo vision and deep learning by computing disparities between learned feature maps rather than raw pixels. In particular, \cite{kendall2017end} use domain knowledge to formulate a novel, differentiable ``soft argmin'' layer for regressing sub-pixel disparity values. Others extend this work by combining the standard disparity loss with an unsupervised learning term, as well as introducing a new learnable argmax operation \cite{smolyanskiy2018importance}. Despite achieving state of the art results on the KITTI autonomous driving benchmark \cite{geiger2012we, geiger2010efficient}, deep stereo methods are still limited by their reliance on ground truth training data derived from LiDAR sensors, which can require careful, manual calibration \cite{geiger2012we} and produce inaccurate labels on some surfaces \cite{tatoglu2012point, yang2011solving}.

\textbf{Self-Supervised Learning.} Recent work at the intersection of robotics and machine learning leverages robot autonomy during data collection. Sources of self-supervision are rich in the robotic manipulation problems that are the focus of our work. For instance, detecting the success of a grasp attempt can be automated using pressure feedback, side cameras \cite{pmlr-v87-jang18a}, or background subtraction schemes \cite{pmlr-v87-kalashnikov18a}. Observation sequences have also be used to learn video prediction tasks in both mobile and stationary robot settings \cite{pathak2018zero, ebert2017self, pmlr-v87-ebert18a}. Although learning from physical interaction can be data intensive \cite{pinto2016supersizing, agrawal2016learning}, self-supervised learning is attractive when data collected for one purpose (e.g., grasping) can be repurposed for another (e.g., video prediction). In this sense, self-supervision can come for free with other robotics tasks, despite its high sample complexity.

\textbf{Uncertainty Aware Deep Learning.} One reason probabilistic models are attractive is that they can quantify the certainty of their own predictions. In a robotics context uncertainty estimates can enable safe decision making \cite{kahn2017uncertainty}. Past work makes a distinction between model uncertainty and aleatoric uncertainty, the latter of which results from inherent noise in the data \cite{kendall2017uncertainties}. Modeling aleatoric uncertainty is particularly useful for depth estimation \cite{kendall2017uncertainties, kendall2018multi} where sensor noise is prevalent.

Our work is, to the best of our knowledge, the first attempt to leverage self-supervised learning via tactile physical interactions toward depth estimation. Because our approach is intended for real world pick-and-place problems, we additionally make use of uncertainty estimation to prevent depth predictions that might result in collisions.

%===============================================================================

\begin{figure}[t!]
  \label{fig:env}
  \centering
  \includegraphics[width=0.45\textwidth]{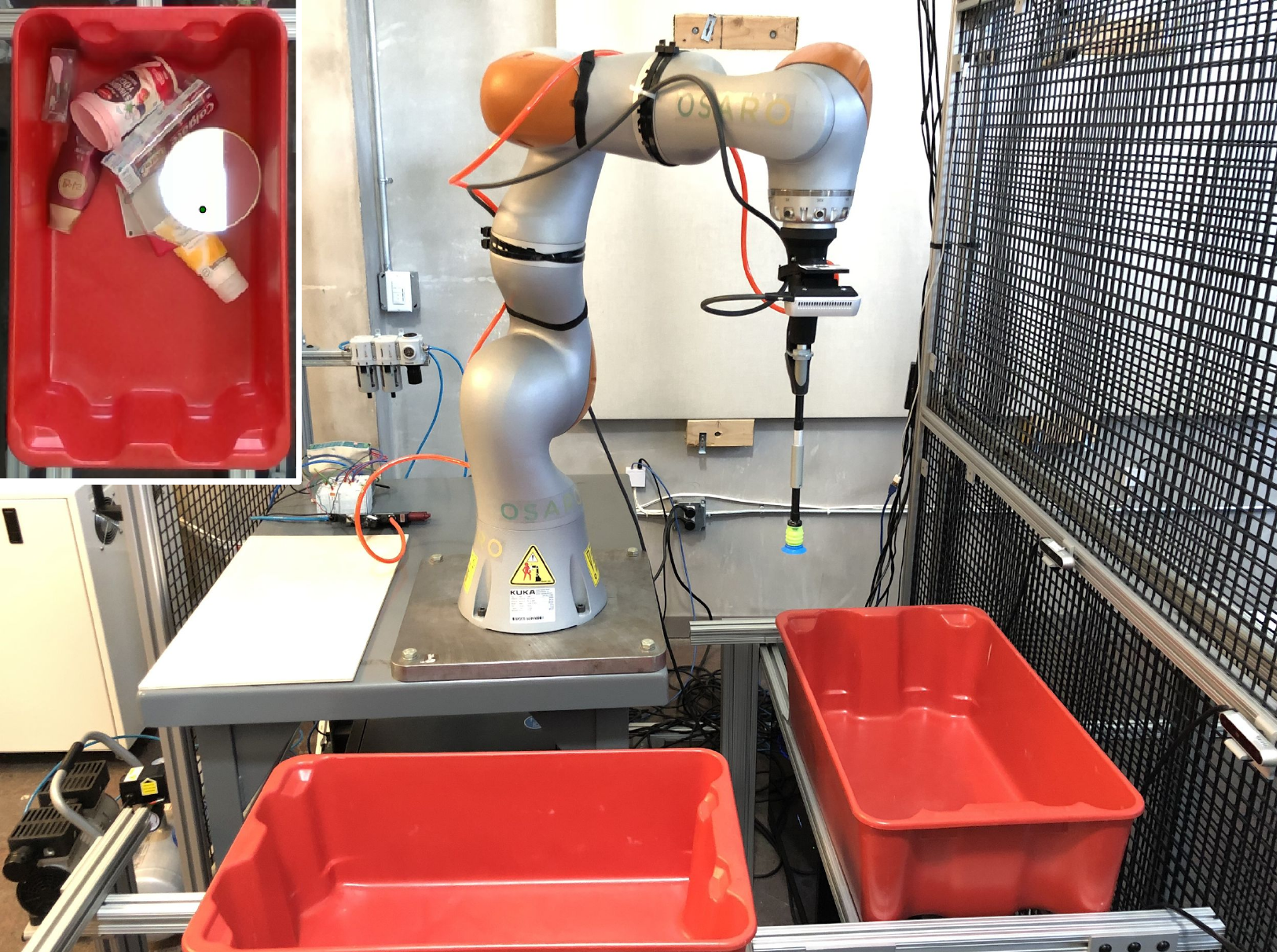}
  \caption{The experimental environment included a Kuka LBR iiwa picking items from a bin. The RealSense camera used for experiments was mounted about the workspace, giving a top-down view into the bin. An example view from the overhead camera is shown in the top left.}
\end{figure}

\section{OVERVIEW}

\label{sec:overview}

\begin{figure}[b!]
  \centering
  \includegraphics[width=0.4\textwidth]{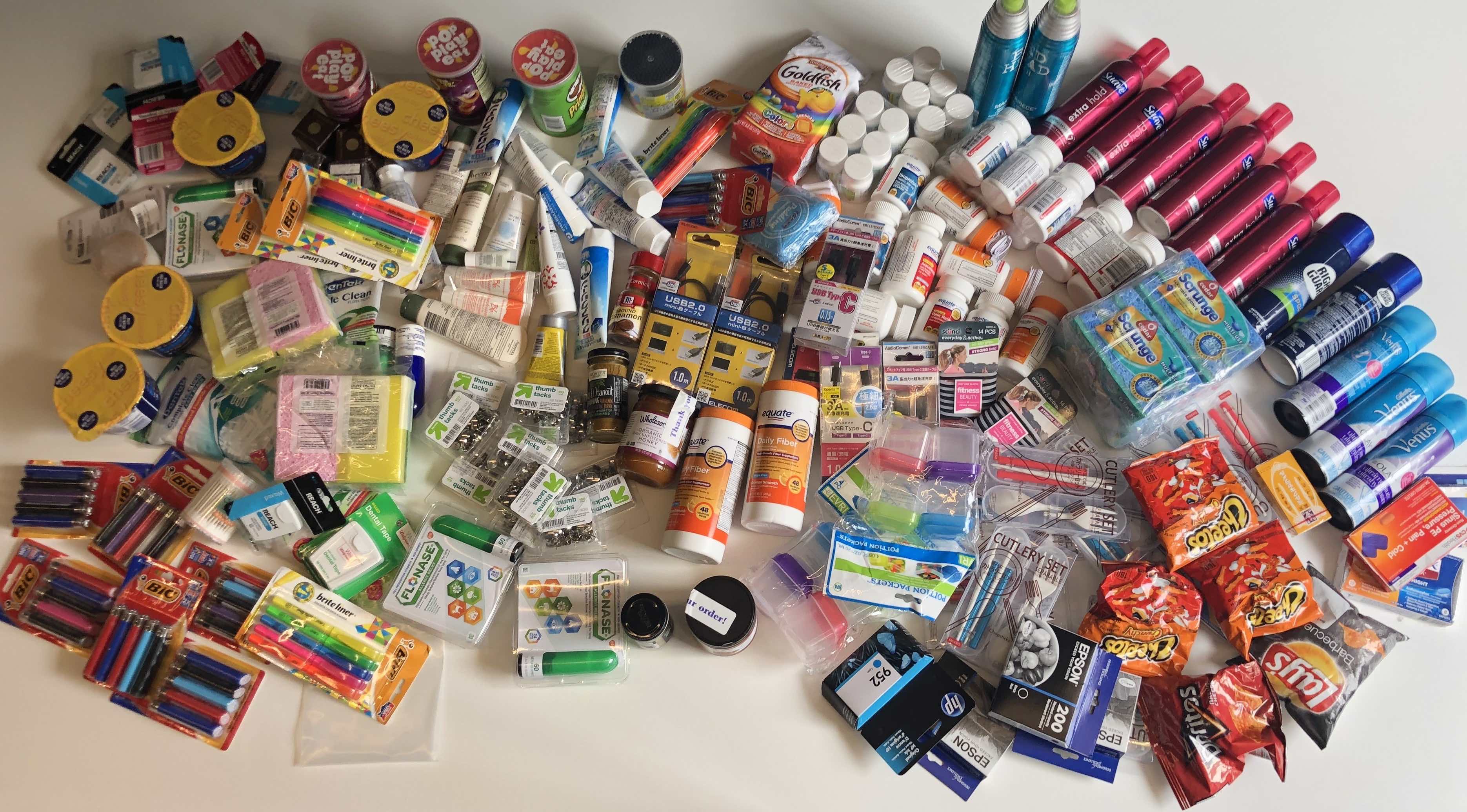}
  \includegraphics[width=0.4\textwidth]{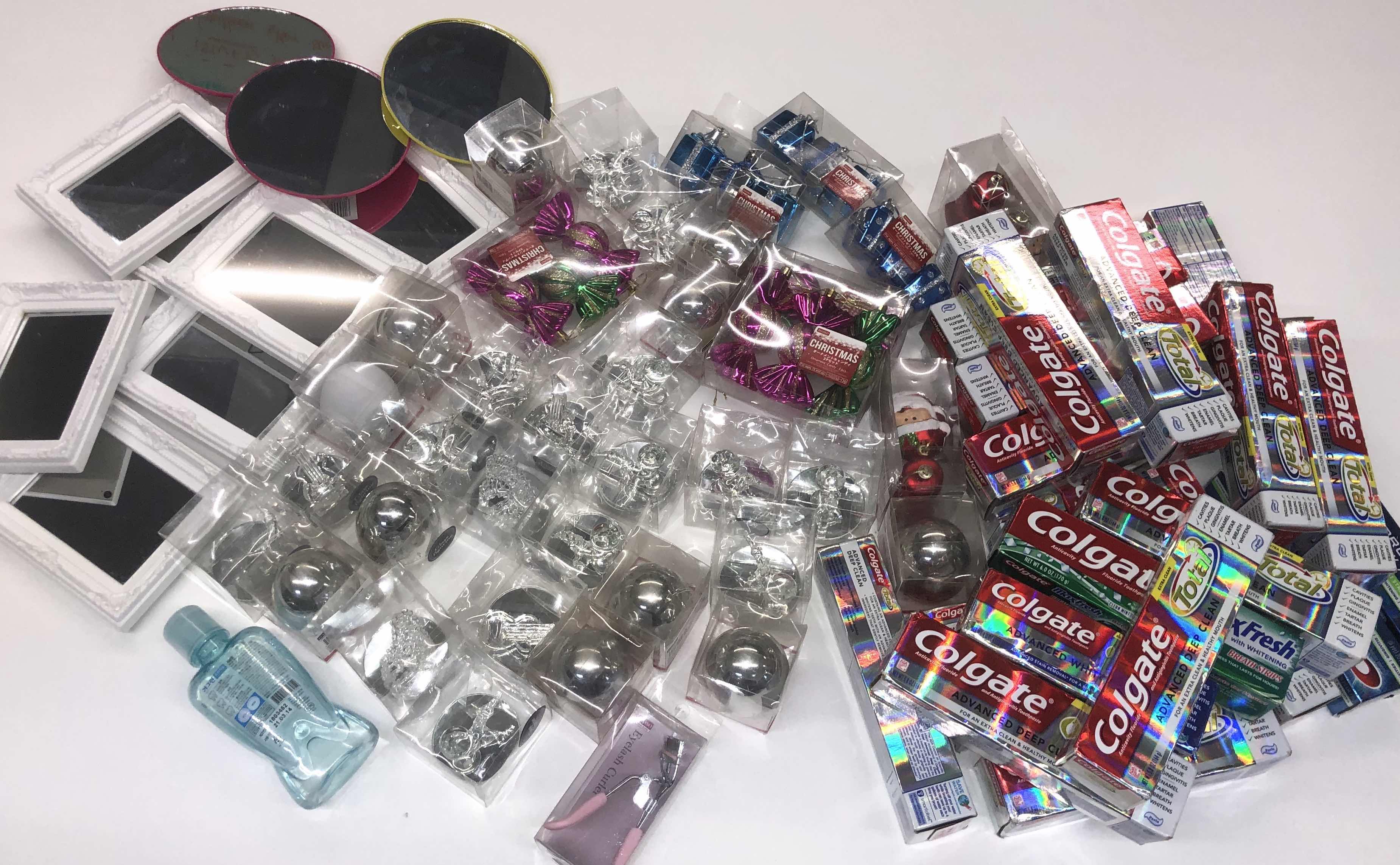}
  \caption{Example items picked in our experiments. Top shows example items from the consumer goods dataset. These include boxes, bottles, aerosels, clamshells, and other items common in retail fulfillment and e-commerce tasks. Bottom shows example adversarial items known to be especially difficult for structured light sensors, such as shiny, transparent, or reflective items.}
  \label{fig:skus}
\end{figure}

\subsection{Problem Formulation}

Our approach assumes a dataset consisting of tuples $(I, D, g, y, z)$ gathered through physical interaction with an environment. Here $I \in \mathbb{R}^{h \times w \times 3}$ is an RGB image and $D \in \mathbb{R}^{h \times w}$ is an aligned depth image sampled from a noisy sensor. This depth image can be viewed as an estimate of $Z \in \mathbb{R}^{h \times w}$, whose values correspond to the actual, ground truth depth at each pixel in $I$.

We wish to produce an estimate $\hat{Z}$ of $Z$ given $I$ and (optionally) $D$. Gathering even a small sample of ground truth depth images is prohibitive for large-scale applications. However, routine data collection and operation produces many single-pixel samples of $Z$, as forward kinematics can recover the tooltip's position along the vertical axis while attempting a grasp. These scalar samples $z$ are recorded at a point $g$ in the image where suction is first achieved or a force threshold is met (for failed grasps). Using camera intrinsic and extrinsic parameters, it is straightforward to map $g$ between world and camera frame. Here we use it to refer to a single pixel of the image $I$. Binary grasp success labels $y$ are also easy to produce in a self-supervised fashion, by relying on suction feedback after a pick attempt.

Given a dataset of attempted picks, we then propose learning $\hat{Z}$ by learning estimates $\hat{z}$ at individual pixels $g$ where picks were attempted. In other words, we formulate depth estimation as a regression problem, where we minimize the distance between $z$ and $\hat{z}$. The labels $y$ can also be used as auxiliary information in our loss.

\subsection{Environment Setup}

Data were collected using a Kuka LBR iiwa 14 R820 fitted with a $34mm$ wide suction tooltip. A RealSense camera mounted directly over the picking bin gathered images $I$ and $D$. Images were sampled at resolutions between $320$ and $512$ pixels in height and width in all experiments. Each arm followed a randomized policy, but our approach is general enough to work with a number of popular grasping methods, such as QT-Opt \cite{pmlr-v87-kalashnikov18a} or DexNet \cite{Mahler2018DexNet3C}.

Robots attempted grasps on well over 100 unique objects representative of stock keeping units (SKU) encountered in real world environments. This object set included, but was not limited to regular boxes, items with reflective or transparent surfaces, clamshells, and aerosol bottles. A sample of SKUs used in our experiments is shown in~\cref{fig:skus}.

\subsection{Model Architecture}
\label{sec:model_architecture}
We formulate depth estimation as an image-to-image translation problem \cite{isola2017image}, where a model with parameters $\theta$ maps input images $I$ and $D$ to an output image $\hat{Z}$. FCNs \cite{long2015fully} work well in this setting, as they preserve the spatial layout of image data while transforming input to output through a series of convolutions and other spatial operations.

\begin{figure}
  \centering
  \includegraphics[width=0.5\textwidth]{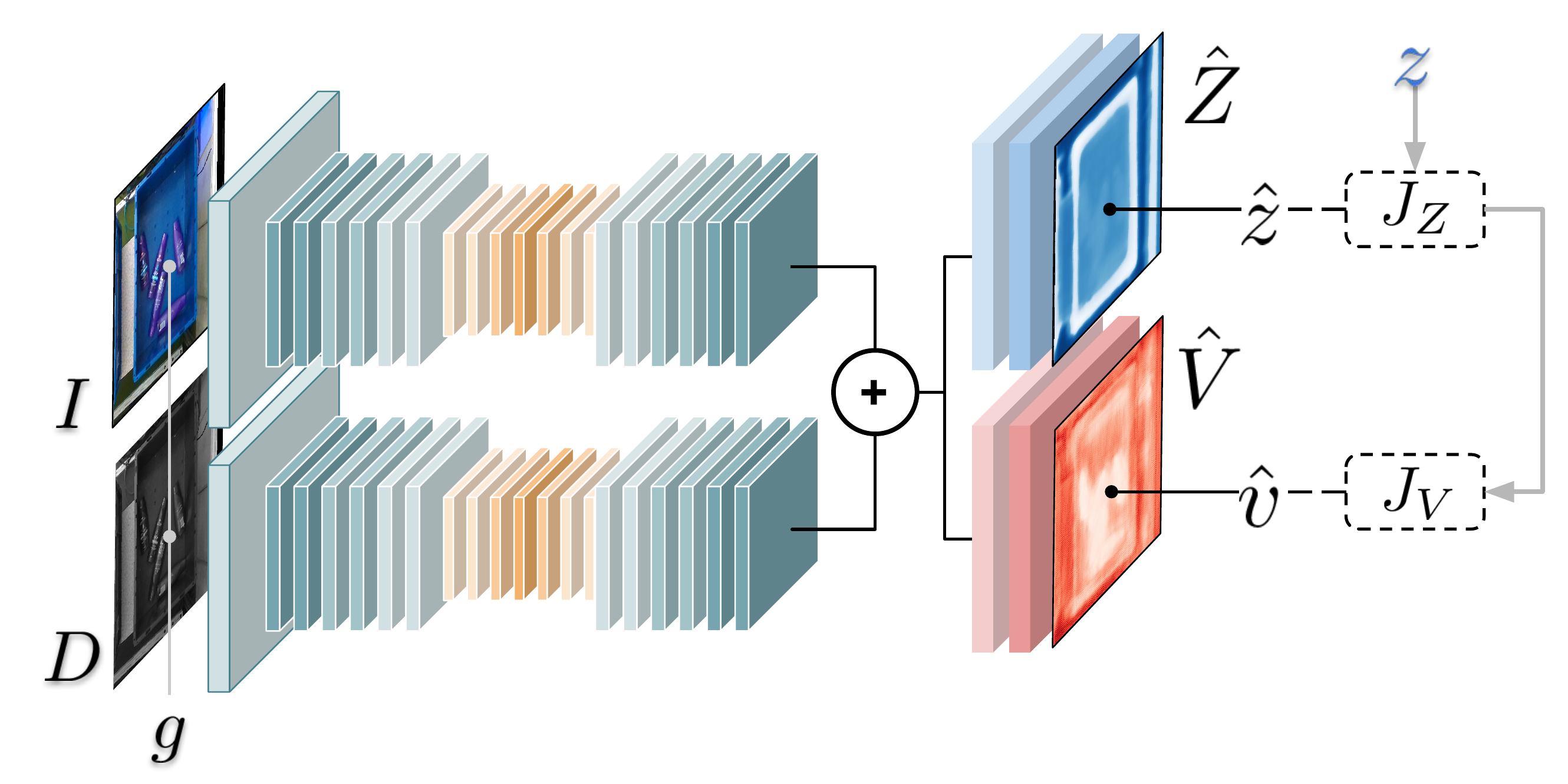}
  \caption{Depth by Poking model architecture. Disjoint feature pyramidal networks are used to encode input images. Feature maps are then processed by a depth estimation head predicting $\hat{Z}$ and an uncertainty estimation head predicting $\hat{V}$. Dashed boxes denote loss functions. Gray arrows denote supervision signal. Elementwise merging of the feature pyramids makes the architecture easy to modify for ``RGB Only'' models.}
  \label{fig:model}
\end{figure}

In our experiments we represent DbP with a FCN having an encoder-decoder architecture with disjoint encoders for the $I$ and $D$ inputs shown in~\cref{fig:model}. The encoders are implemented as feature pyramidal networks (FPN) \cite{lin2017feature}, each with a ResNet-101 backbone \cite{he2016deep} pretrained on the MS COCO object detection dataset \cite{lin2014microsoft}. Since the depth encoder has one channel, its input layer can't be loaded from COCO and is instead randomly initialized. The encoders' output feature maps are merged by elementwise addition. The decoder is a simpler architecture that uses convolution transpose layers to upsample feature maps followed by dimension-preserving convolution layers with $3 \times 3$ filters. The output layer has linear activations and $1 \times 1$ convolutions.

\begin{figure}[b!]
  \centering
  \includegraphics[width=0.4\textwidth]{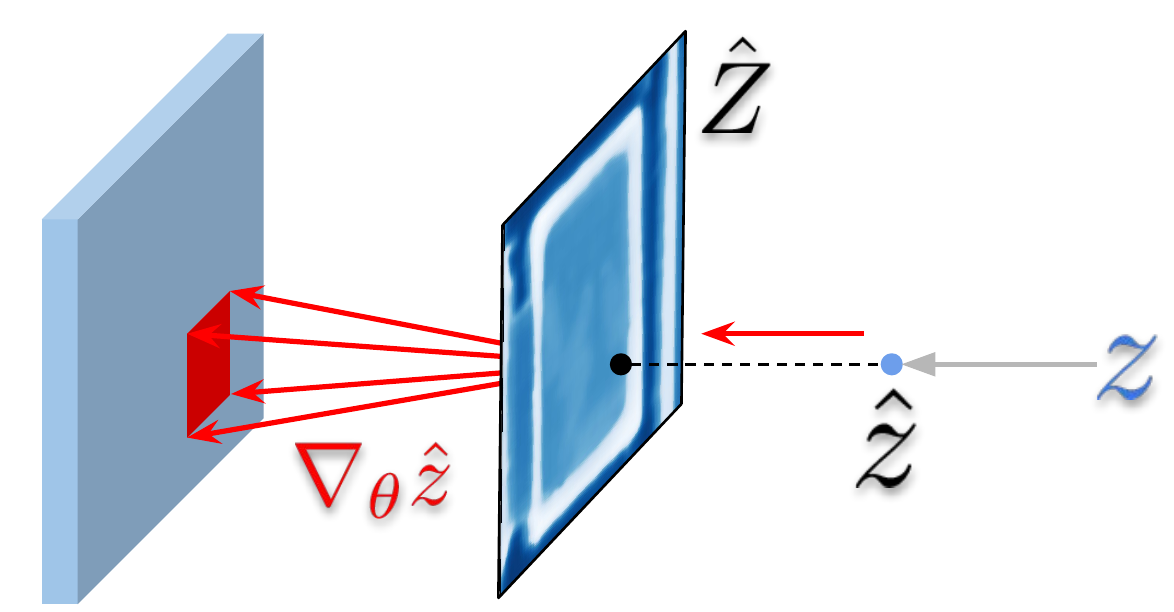}
  \caption{Pixelwise training. Loss is computed at one pixel, indexed by $g$, per sample. Red arrows denote gradient flow. Gray arrow denotes supervision signal.}
  \label{fig:pixelwise}
\end{figure}

\section{Depth by Poking}
\label{sec:dbp}

We introduce a self-supervised depth estimation system learned by attempted grasps, or ``pokes'' in a physical environment. Depth by Poking (DbP) attempts to solve two problems. First, we would like accurate depth estimates. Second, we would like to quantify our certainty in these predictions. We describe differentiable objectives for achieving these goals in the following subsections.

\subsection{Depth Estimation}
\label{sec:depth_estimation}

We wish to minimize our loss on a pixelwise basis. Because each $\hat{z}$ estimates a continuous value $z$, minimizing mean squared error (MSE) is appropriate. The squared error for the $i^{th}$ training example is given by:
\begin{equation}
\label{eq:residual}
    v_i = (\hat{z}_i - z_i)^2
\end{equation}

Minimizing the average of~\cref{eq:residual} with respect to $\theta$ over the entire dataset would treat all grasp points equally. However, its reasonable to assume that unsuccessful grasps have noisier depth labels than successful grasps (e.g., failure may have caused the arm to retract before reaching the grasp point). Therefore, we favor a loss that assigns different weights to samples with positive or negative labels:
\begin{equation}
    \label{eq:depth}
    J_Z \\ = \frac{\lambda_+}{N_+} \sum_{i=1}^{N} y_i v_i + \frac{\lambda_-}{N_-} \sum_{i=1}^{N} (1-y_i) v_i
\end{equation}

Where $N_+ = \sum_{i=1} y_i$ and $N_- = \sum_{i=1} (1-y_i)$ and $\lambda_+$ and $\lambda_-$ are the weights for the successful and unsuccessful grasps respectively.

These losses only accumulate over pixels of the output map $\hat{Z}$ for which the grasp points $g$ exist. As visualized in~\cref{fig:pixelwise}, learning signal can still propagate into larger portions of the network through the indexed activation's receptive fields.

\subsection{Uncertainty Estimation}

Accurate depth estimation is vital for pick-and-place, as we rely on depth maps to determine where the gripper should go. Underestimating the depth causes the effector to stop short of its destination, meaning the motion must be completed with slow, tactile feedback. Overestimating the depth can cause a hard collision. In addition to estimating pixelwise depth, we therefore wish to estimate our uncertainty in our estimates to avoid unsafe actions.

Uncertainty can be divided into two broad categories: epistemic model uncertainty resulting from limited training data and aleatoric uncertainty resulting from sensor noise \cite{kendall2017uncertainties}. In the self-supervised industrial pick-and-place setting we assume data are abundant. Sensor noise is however prevalent as depth sensing hardware produces poor estimates for reflective, transparent, or shiny surfaces. We therefore wish to model aleatoric uncertainty.

Modeling aleatoric uncertainty requires minor changes to our architecture and loss function, but training and data collection otherwise remain the same. First, we add an additional output map: A variance map $\hat{V}$ with the same height and width as $\hat{Z}$, $I$ and $D$. Each pixel of $\hat{V}$ stores the estimated variance, or squared error $\hat{v}$ for the corresponding depth estimate in $\hat{Z}$. In order to learn $V$ we explore two alternative training approaches.

\subsubsection{Gaussian Log Likelihood Uncertainty Estimation}
Assuming depth values $z$ are sampled from a normal distribution with mean $\mu$ and variance $v$ conditioned on some input, we can estimate aleatoric uncertainty through maximum likelihood estimation \cite{kendall2017uncertainties, kendall2018multi}.
\begin{equation}
    \label{eq:loglike}
    J_\mathcal{N} = \frac{1}{N} \sum_{i=1}^{N} \frac{1}{2 \hat{v}_{i}}(z_i - \hat{z}_i)^2 + \frac{1}{2}\log \hat{v}_{i}
\end{equation}

Here $\hat{z}_i$ and $\hat{v}_{i}$ are estimates for $\mu$ and $\sigma$ at grasp pixel $g$. Both variables depend on model inputs and parameters. Backpropagating through these estimates yields an objective that attempts to minimize MSE but can shrink loss on uncertain examples (i.e., where $\hat{v}_{i}$ is large). We refer to models trained on this objective as ``\LLDbP''.

\subsubsection{Moments-Based Uncertainty Estimation}

Alternatively, we can add an additional loss term for regressing the values of $\hat{V}$ directly. Here we make use of the fact that the average of residuals in~\cref{eq:residual} approximates the variance of the data conditioned on the model inputs \cite{bishop1994mixture}. We can then use the residuals themselves as labels and minimize the objective:
\begin{equation}
    \label{eq:variance}
    J_V = \frac{1}{N} \sum_{i=1}^{N} (v_i \stopgrad - \hat{v}_i)^2 \\ = \frac{1}{N} \sum_{i=1}^{N} ((\hat{z}_i \stopgrad - z_i)^2 - \hat{v}_i)^2
\end{equation}

Similarly to~\cref{eq:loglike}, $\hat{v}$ denotes the value of $\hat{V}$ sampled at grasp point $g$. We use $\cdot \stopgrad$ to denote the stop gradient operation, which prevents any gradients with respect to $J_V$ from updating the parameters of the depth estimation head during training. This operation prevents large errors predicting $v$ from affecting predictions of $z$ during training.

\begin{figure}[b]
  \centering
  \includegraphics[width=0.5\textwidth]{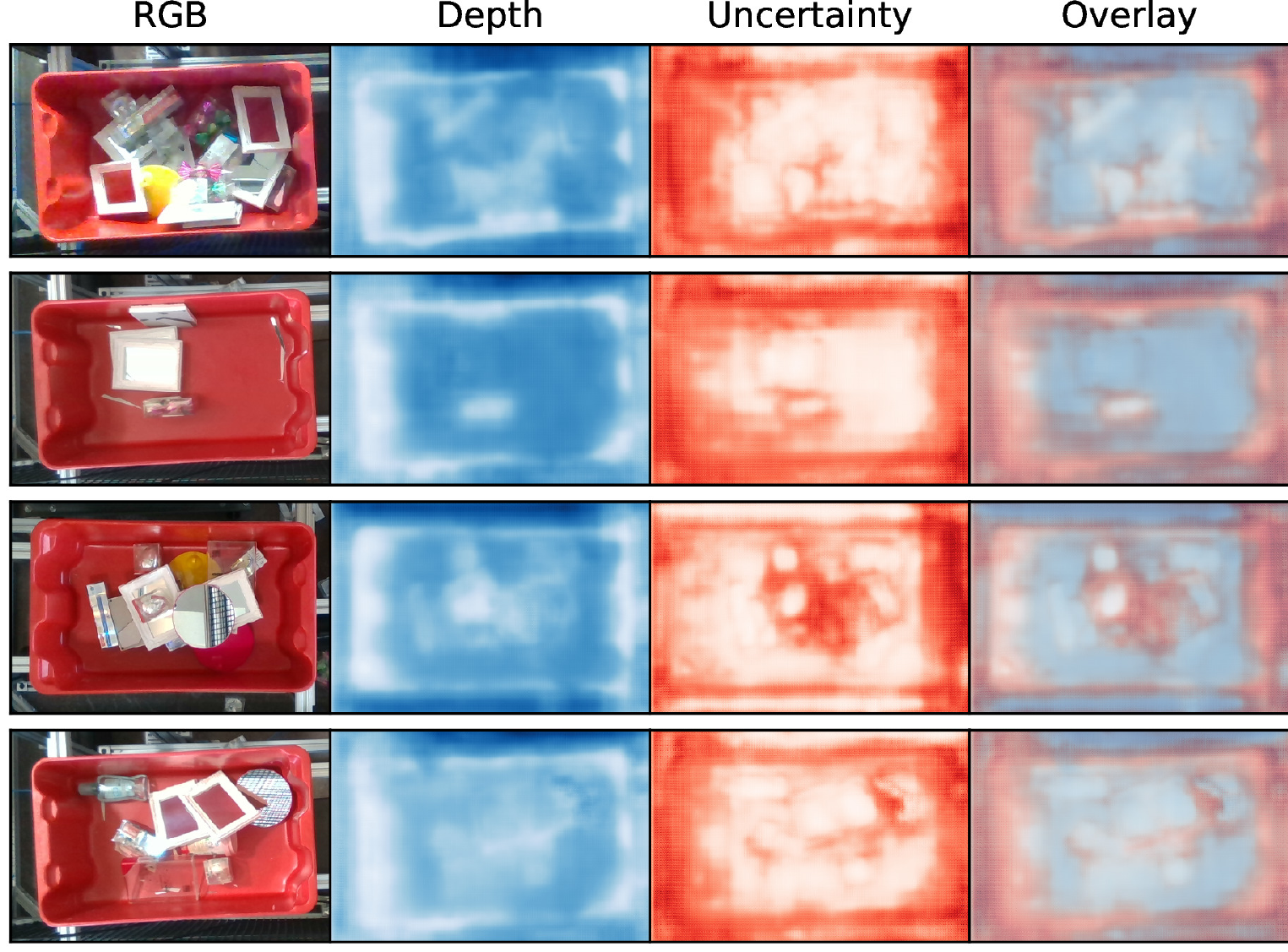}
  \caption{Predictions on an example image. Lighter pixels denote shorter distance to the camera in the predicted depth image $\hat{Z}$ and lower predicted variance in the uncertainty image $\hat{V}$. The final frame overlays the two previous, visualizing the relationship between depth and uncertainty. In general we see the model is more certain around pickable objects, where the data coverage is greatest. Best viewed in color.}
  \label{fig:qualitative}
\end{figure}

One benefit of optimizing~\cref{eq:variance} is that it estimates $V$ separately from $Z$. We can therefore weight our loss by $\lambda_{V}$ depending on how much uncertainty estimation matters to a given application. Combining~\cref{eq:depth,eq:variance}, we have the objective function:
\begin{equation}
    \label{eq:dbp}
    J_{M} = J_Z + \lambda_{V} J_V
\end{equation}

Minimizing~\cref{eq:dbp} with respect to $\theta$ then learns a model that both estimates depth and its aleatoric uncertainty on a pixelwise basis. We refer to such models as ``\MDbP''.

%===============================================================================

\section{EXPERIMENTS}

Through our experiments we sought to answer three main questions about our learning method. First, how well does DbP estimate the depth of attempted picks in a realistic, industrial application? Second, how robust is our system to items and surfaces upon which standard methods fail? And third, can estimated uncertainty help us mitigate problems caused by poor depth estimates?

\begin{figure*}[h!]
  \centering
  \includegraphics[width=1.0\textwidth]{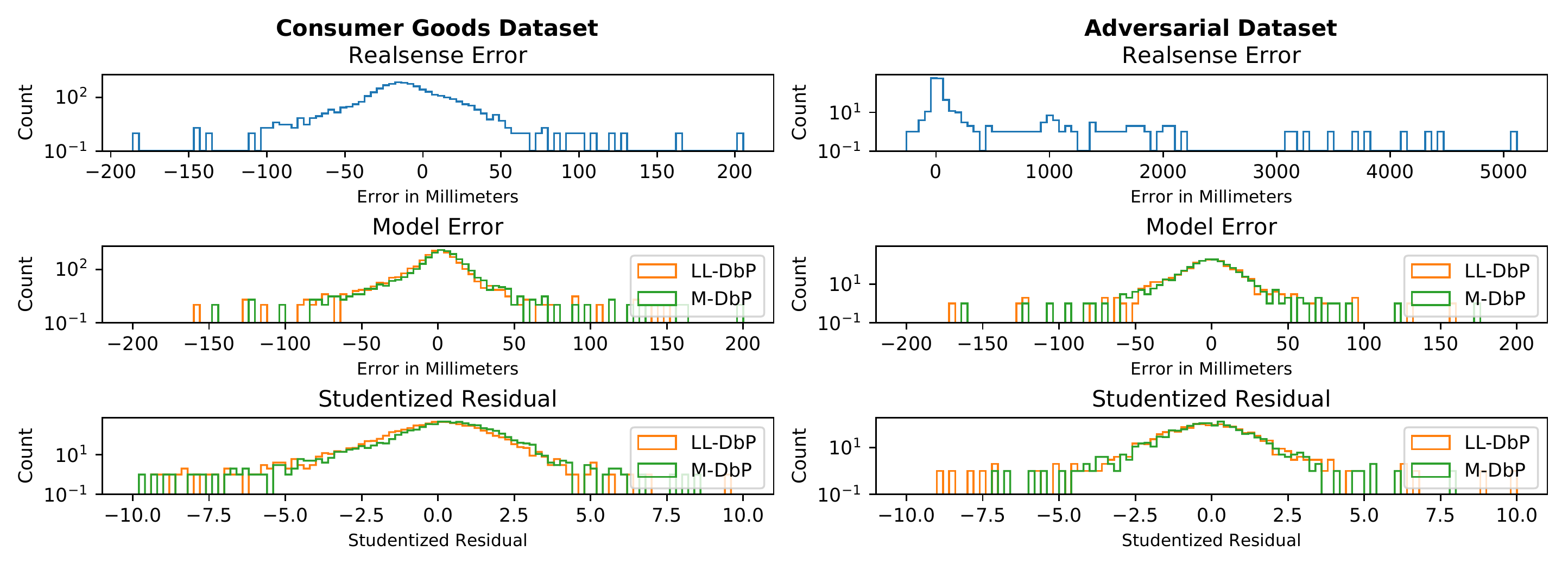}
  \caption{Log-scale histograms showing raw RealSense error (top), the error distribution of our model (middle), and the error distribution of our model divided by the predicted standard deviation (bottom). Left is the consumer goods dataset and right is the adversarial items dataset.}
  \label{fig:histogram}
\end{figure*}

\subsection{Datasets}

We collected two datasets using the Kuka iiwa. First, the ``consumer goods'' dataset consists of 58,168 picks attempted on items in jumbled bins at various levels of fullness. The objects themselves were drawn from a set of well over 100 household and food items common to real-world warehousing, e-commerce, and retail fulfillment settings. Second, the ``adversarial dataset'' also consisted of top-down images of jumbled bins. However in this setting items were chosen to be difficult for traditional depth estimation methods. Such items include transparent glassware and plastic, mirrors and objects with reflective of shiny surfaces. This dataset consisted of 14,934 attempted picks. See~\cref{fig:skus} for representative samples of SKUs in each dataset.

Data were collected autonomously in both settings. The robot attempted grasps on items in a pick-bin and transported them to a place-bin in the event of a success. Once the pick-bin was cleared, the policy continued by treating the place bin as the pick bin and vice versa. An operator intervened only every few hours to reset the bins with new items. This procedure allowed us to gather data at various bin states including full, sparse, and nearly empty.

One limitation of self-supervised learning and data collection at this scale is dealing with noise caused by hardware failure. Force sensors occasionally malfunction, which can trigger early retraction and thus inaccurate depth labels. We manually found and discarded many of these bad datapoints where possible, as the purpose of our study was to evaluate depth estimation methods, rather than the hardware or picking policy. Nevertheless, we expect some label noise in real-world settings, which motivates the use of uncertainty estimation and differential weighting of succeeded and failed picks in our loss through hyperparameters $\lambda_+$ and $\lambda_-$. After post-processing, both datasets were split into training and test sets according to a $9:1$ split. We report results over the test set in the following experiments.

\subsection{Baselines}
\label{sec:baselines}
We compared our approach to two main types of baseline depth estimators. Because the focus of our work is on autonomous depth estimation without human annotation, appropriate baselines are either self-supervised, unsupervised, or not learned from data at all.

\subsubsection{Structured Light Sensors} An Intel RealSense D415 structured light sensor (SLS) is used to produce depth maps. We experimented with Gaussian filtering of the estimated depth maps and report the baseline obtaining the lowest root mean squared error (RMSE). Due in part to the non-rigid suction cup compressing before pressure feedback is detected, there may be some constant bias between the depth map and the position at which the robot stops. DbP models won't be affected by this distance, as they are trained directly on samples from $Z$, rather than $D$. However, comparing learned methods and SLS's naively fails to account for the bias. Therefore we computed the bias empirically and subtracted the resulting constant from our measurements.

\subsubsection{Deep Autoencoders} Convolutional neural networks are known to provide useful inductive biases on many vision tasks, as demonstrated by the surprising performance of even untrained networks~\cite{saxe2011random}. We therefore questioned whether the performance of DbP was due to its training criterion or simply its architecture. Our second family of baselines answers this question by training the architecture described in~\cref{sec:model_architecture} according to an L2 reconstruction loss on the RealSense depth image, rather than~\cref{eq:dbp}.

\begin{table}[t]
\caption{RMSE between predicted and ground truth tooltip depth, reported over both test sets. All figures are in millimeters.}
\begin{center}
\begin{tabular}{ccc} \toprule
     & Consumer Good & Adversarial \\ \midrule
    Dataset Size & 58,168 & 14,934 \\ \midrule
    RealSense Raw & 22.06 & 427.89 \\
    RealSense BC & 18.33 & 418.43 \\
    RealSense GF & 18.65 & 154.06 \\
    RealSense GF/BC & 17.80 & 145.41 \\
    Autoencoder & 22.14 & 228.07 \\
    Autoencoder BC & 18.57 & 220.81 \\ \midrule
    \MDbP (RGB Only) & 18.84 $\pm$ 0.52 & 23.11 $\pm$ 1.07 \\
    \LLDbP (RGB Only) & 19.46 $\pm$ 0.70 & 23.67 $\pm$ 0.42 \\
    \textbf{\MDbP (RGB-D)} & 13.09 $\pm$ 0.13 & 20.48 $\pm$ 1.60 \\
    \textbf{\LLDbP (RGB-D)} & 13.25 $\pm$ 0.33 & 20.05 $\pm$ 0.98 \\
    \bottomrule
\end{tabular}
\end{center}
\label{table:rmse_results}
\end{table}

\subsection{Depth Estimation}

RMSE results in millimeters are reported in~\cref{table:rmse_results}. Several RealSense baselines are included along with two versions of the autoencoder baseline. ``BC'' denotes models featuring the bias correction described in~\cref{sec:baselines}. ``GF'' denotes Gaussian filtered depth maps, using kernel width $\sigma = 13$. We compared multiple filter widths and report the baselines obtaining the best results. RealSense ``Raw'' has no post-processing applied to its predicted depth maps beyond those implemented by the camera firmware.

DbP can be interpreted as either a monocular depth estimation approach or a depth-denoiser, depending on whether the model conditions on $D$. We test both (RGB Only vs. RGB-D) in our experiments. As the RGB-D models performed best, we trained multiple initializations on both datasets to establish the standard deviations shown in~\cref{table:rmse_results}. All other hyperparameters were fixed.

For the consumer goods dataset, the best overall baseline is RealSense GF/BC at $17.8mm$, which is substantially larger than the errors obtained by DbP. Our best model on this dataset, \LLDbP, achieved an RMSE of $13.09mm$ with a standard deviation of 0.13.

For the adversarial dataset, all baselines failed to predict reasonable estimates at grasp points. Inaccurate values reported by the RealSense camera caused these errors, even in the autoencoder baseline which relied on noisy depth maps for training data. For example, we observed that depth predictions of the RealSense camera to a mirror's surface could be off by up to 5 meters. These large errors can be seen in the top right of~\cref{fig:histogram}. DbP avoids these errors by predicting end effector positions, which are not sensitive to surface specular properties. Nevertheless, the adversarial items still proved more difficult for DbP than the consumer goods dataset, even though it consisted of fewer items. Here the RGB Only models performed more similarly to the RGB-D models, indicating that DbP can use visual cues, rather than simply denoising the depth input.

In order to understand the distribution of depth estimation errors, we also show RMSE histograms in~\cref{fig:histogram}. The right and left columns correspond to the consumer and the adversarial datasets respectively. The top histogram shows raw RealSense errors. Positive values on the histogram correspond to over-estimates, where $\hat{z}$ exceeded $z$. Negative values correspond to under-estimates. The bias due to gripper compliance can be seen in the top left histogram where the peak of the distribution is slightly shifted left. Here the suction cup deforms as it reaches a centimeter or two beyond the predicted depth. The second row corresponds to RMSEs of DbP, which are zero centered with no bias. The bottom row shows the model error divided by the model's predicted standard deviation, corresponding to a studentized residual \cite{cook1982residuals}. If the model correctly predicts a mean and standard deviation of a normal distribution, then we'd expect the studentized residuals to be unit normal. In general this is what we see, except for a few large outliers, for which noise in the self-supervised labels appear to be the cause.

\begin{figure}[t!]
  \centering
  \includegraphics[width=0.5\textwidth]{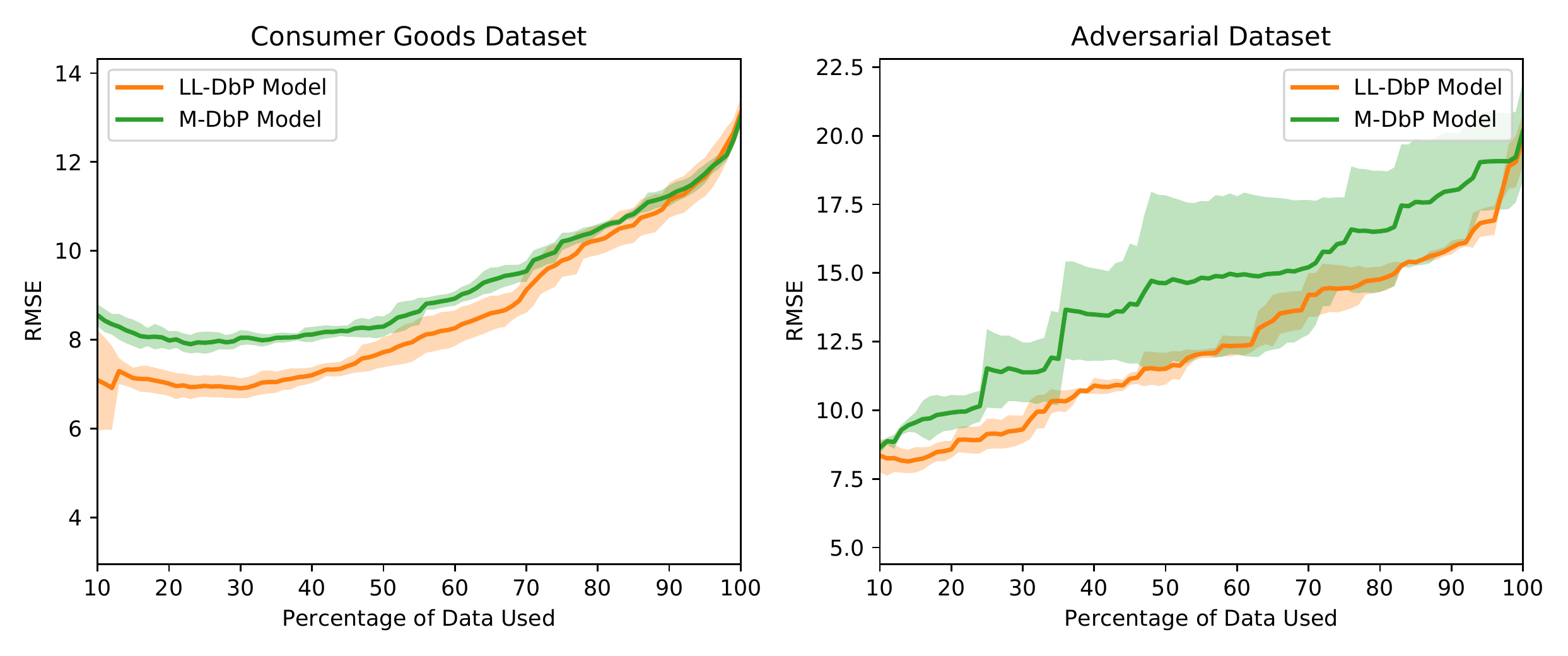}
  \caption{Data discard plots illustrating how RMSE decreases as data are removed based on the percentile of the predicted variance for that datapoint. The curve is averaged over several models trained from different initializations. The shaded region shows 1 standard deviation.}
  \label{fig:datadiscard}
\end{figure}

\subsection{Uncertainty Estimation}

In real-world applications, one use case of uncertainty estimation is to prune potentially incorrect depth estimates, so that we may rely only on those in which we are confident. Therefore, we quantify uncertainty estimation in DbP by measuring RMSE of the depth predictions for various values of uncertainty, as has been done in prior work on uncertainty aware deep learning \cite{kendall2017uncertainties}. In other words, we rank picks in the test set by predicted uncertainty, and show how the RMSE of depth estimates changes when we evaluate the model at various percentiles. Results are shown in~\cref{fig:datadiscard}. If the uncertainty accurately quantifies the data's variance, then the overall RMSE should tend to decrease as the data points with the largest predicted variance are removed.

\begin{figure}[t!]
  \centering
  \includegraphics[width=0.5\textwidth]{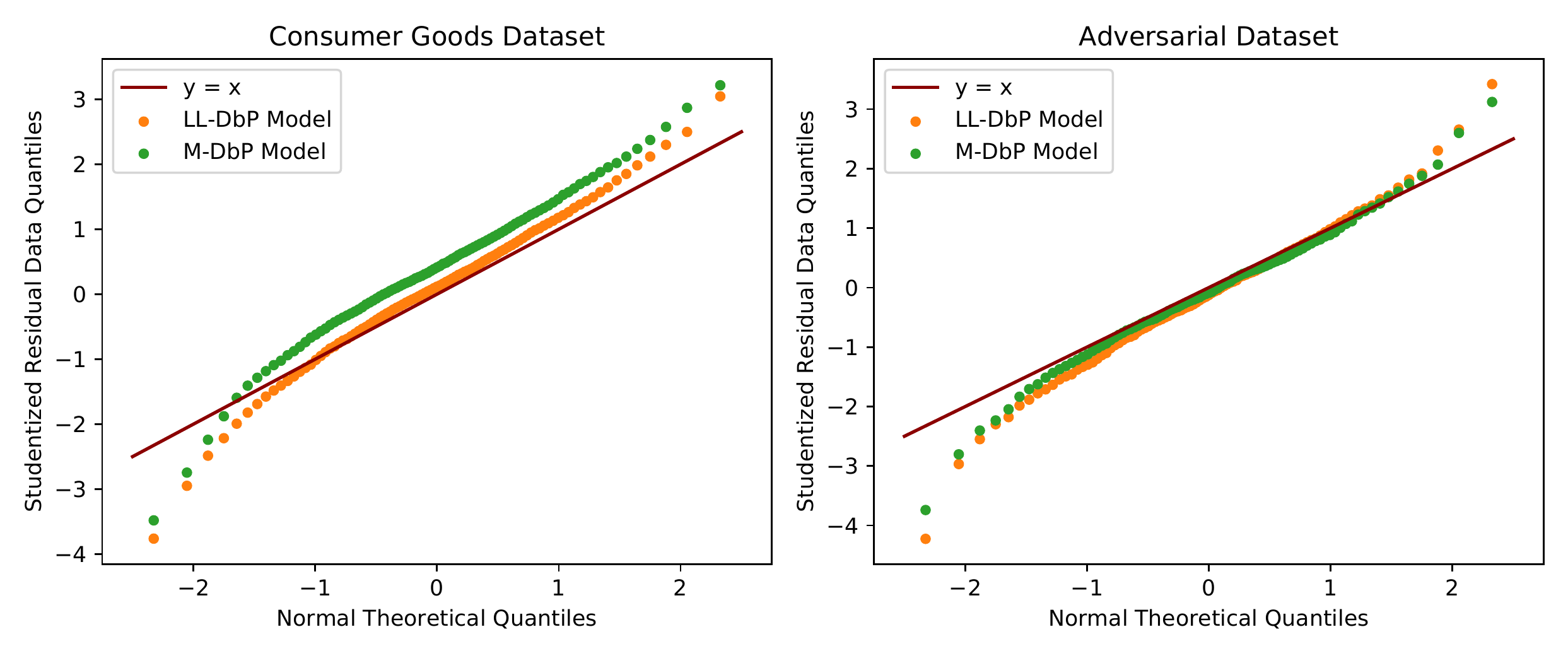}
  \caption{Q-Q plots showing the quantiles of the studentized residual compared against Gaussian quantiles.}
  \label{fig:qq}
\end{figure}

Since this model is trained to produce errors under a Gaussian Likelihood assumption, we expect the studentized residuals to correspond to a standard normal distribution. The Q-Q plots in~\cref{fig:qq} reveal that this expectation generally holds. However there are some large outliers both for over and under predictions, some of which are caused by data mislabeled during the self-supervised collection process.

Finally we visualize some qualitative results in~\cref{fig:qualitative}. Here we observe less certainty near the edges of objects and in regions where items are stacked. These predictions likely model the slightly random behavior of an end effector attempting grasps in such areas. Stacks may shift under pressure and items are likely to slide when grasped at the edges, producing aleatoric uncertainty. The model also predicts the depth of mirror surfaces with little uncertainty.

\section{CONCLUSIONS}
Depth by Poking improves on standard structured light based depth estimation by training a model on physical pressure feedback measurements from an end effector. We've shown this technique generalizes to complete depth maps, despite being trained on single-pixel labels. DbP can also model uncertainty with minor changes to its architecture and loss function. Here we experimented with two types of uncertainty representations and demonstrate that both were helpful for pruning hard picks. This approach outperforms structured light sensors and deep autoencoder baselines for both non-reflective objects, as well as reflective, shiny, or transparent objects on which other methods fail.

%                                   % the textheight of the last page by a suitable amount.
%                                   % This command does not take effect until the next page
%                                   % so it should come on the page before the last. Make
%                                   % sure that you do not shorten the textheight too much.

%%%%%%%%%%%%%%%%%%%%%%%%%%%%%%%%%%%%%%%%%%%%%%%%%%%%%%%%%%%%%%%%%%%%%%%%%%%%%%%%

%%%%%%%%%%%%%%%%%%%%%%%%%%%%%%%%%%%%%%%%%%%%%%%%%%%%%%%%%%%%%%%%%%%%%%%%%%%%%%%%



\section*{ACKNOWLEDGMENT}
We would like to thank our colleagues at Osaro including, Khashayar Rohanimanesh, Rishi Sharma, Chris Correa, Sulabh Kumra, Eddie Groshev, Kazu Komoto, Sebastiaan Boer, Sash Nagarkar, and many others on the robotics control and infrastructure teams. Their work made this project possible.

%%%%%%%%%%%%%%%%%%%%%%%%%%%%%%%%%%%%%%%%%%%%%%%%%%%%%%%%%%%%%%%%%%%%%%%%%%%%%%%%

\bibliographystyle{IEEEtran}
\bibliography{IEEEabrv,root}

\end{document}